\documentclass{article}

    \usepackage[preprint]{neurips_2020}

\usepackage[utf8]{inputenc} 
\usepackage[T1]{fontenc}    
\usepackage{hyperref}       
\usepackage{url}            
\usepackage{booktabs}       
\usepackage{amsfonts}       
\usepackage{nicefrac}       
\usepackage{microtype}      
\usepackage[pdftex]{graphicx}
\usepackage{amsmath}
\usepackage{multirow} 
\usepackage{bm}
\usepackage{subfigure}
\usepackage{appendix}
\usepackage[inline]{enumitem}
\usepackage{wrapfig}

\newtheorem{case}{Case}

\title{Inverse Graph Identification: Can We Identify \\ Node Labels Given Graph Labels?}

\author{
  Tian Bian\\
  Tsinghua University\\
  \texttt{bt18@mails.tsinghua.edu.cn}\\
  \AND
  Xi Xiao$^*$\\
  Tsinghua University\\
  \texttt{xiaox@sz.tsinghua.edu.cn}\\
  \And
  Tingyang Xu\\
  Tencent AI Lab\\
  \texttt{tingyangxu@tencent.com}\\
  \And
  Yu Rong\\
  Tencent AI Lab\\
  \texttt{yu.rong@hotmail.com}\\
  \And
  Wenbing Huang\thanks{Co-corresponding authors.}\\
  Tsinghua University\\
  \texttt{hwenbing@126.com}\\
  \And
  Peilin Zhao\\
  Tencent AI Lab\\
  \texttt{masonzhao@tencent.com}\\
  \And
  Junzhou Huang\\
  Tencent AI Lab\\
  \texttt{joehhuang@tencent.com} \\
}

\begin{document}

\maketitle

\begin{abstract}

Graph Identification (GI) has long been researched in graph learning and is essential in certain applications (\emph{e.g.} social community detection). Specifically, GI requires to predict the label/score of a target graph given its collection of node features and edge connections. While this task is common, more complex cases arise in practice---we are supposed to do the inverse thing by, for example, grouping similar users in a social network given the labels of different communities. This triggers an interesting thought: can we identify nodes given the labels of the graphs they belong to? Therefore, this paper defines a novel problem dubbed Inverse Graph Identification (IGI), as opposed to GI. 
Upon a formal discussion of the variants of IGI, we choose a particular case study of node clustering by making use of the graph labels and node features, with an assistance of a hierarchical graph that further characterizes the connections between different graphs. To address this task, we propose Gaussian Mixture Graph Convolutional Network (GMGCN), a simple yet effective method that makes the node-level message passing process using Graph Attention Network (GAT) under the protocol of GI and then infers the category of each node via a Gaussian Mixture Layer (GML). The training of GMGCN is further boosted by a proposed consensus loss to take advantage of the structure of the hierarchical graph. Extensive experiments are conducted to test the rationality of the formulation of IGI. We verify the superiority of the proposed method compared to other baselines on several benchmarks we have built up. We will release our codes along with the benchmark data to facilitate more research attention to the IGI problem.

\end{abstract}

\section{Introduction}
In many scenarios, the objects with their features are connected by their interactions as graphs. By analyzing these edge connections and node features throughout various graphs, a graph identification (GI) problem is formed to predict the information or properties of the graphs, such as labels or scores of the target graphs. Many studies have been developed on GI, such as graph classification~\cite{li2019semi,shi2000normalized} and graph regression~\cite{pujara2013knowledge}. Formally, methods for GI aggregates the information from nodes and edges to predict or summarize the information of the whole graph. However, it is much more interesting to consider an inverse problem which has never been proposed: Can we use the information of graphs to infer the information of nodes or even edges and sub-graphs? In another word, given the labels of graphs, how to figure out the categories of nodes, edges, or sub-graphs? 

This problem is interesting and very common in the real world. In social media, for example, thinking of hot events that are widely discussed in the form of graphs where users involved as nodes and the co-following relationship among users as edges, it is interesting to think how to pick out the malicious users from the mass of users with only the labels of these event topics such as the authenticity of each topic~\cite{zhou2018fake,cao2018automatic}. In drug discovery, for another example, thinking of molecules as graphs in which atoms as nodes and chemical bonds as edges, we attempt to identify the roles of certain sub-graphs, or equivalently functional groups, in each molecular given its chemical or physical properties~\cite{gilmer2017neural,ma2020dual}. In a programming language, for the last example, thinking of control flows for programs as graphs by considering statements as nodes and control flows as edges, can we detect the problematic statements if we have already known which program has bug or not~\cite{lu2005bugbench}?

All these problems can be defined as a general problem as {\em Inverse Graph Identification} (IGI) that identifies the nodes in graphs based on the information of graphs. The main difficulty of the IGI task is that the node labels are inaccessible so that all available information for the training model only comes from the labels of graphs. Namely, it is a node identification task where the available training labels are much coarser-grained than the node labels we want to fit. 
It seems this problem can be resolved from the present perspective of node clustering~\cite{girvan2002community} or node identification~\cite{kipf2017semi}. 
However, they are different from the problems that we attempt to address. Unlike node clustering which only conducts clustering based on node features or graph structures, IGI is better guaranteed by the given label information of graphs, which, we will demonstrate, closely influences the clustering results. It is also different from the node identification task because IGI does not contain any node labels for training. Another similar concept to our IGI is Multiple-Instance Learning (MIL)~\cite{carbonneau2018multiple} that adopts global labels of bags to identify the labels of local instances. However, MIL assumes that each instance is i.i.d. and there are no edge connections involved. Therefore, it is interesting and important to study the IGI problem as a set of new challenges and seek new solutions for it.

In this paper, we formally define IGI and address out its different problem statements as a set of new challenges. Meanwhile, we focus on a particular study of IGI: A node clustering task by making use of graph labels and node features with an assistance of a hierarchical graph~\cite{girvan2002community} that further characterizes the relations among graphs. To address this particular task, we propose a novel model based on Gaussian mixture model (GMM) and graph convolutional network (GCN) named as Gaussian Mixture Graph Convolutional Network (GMGCN). First, the features of each node are updated through the Graph Attention Network (GAT). Then the node features are aggregated by a Gaussian mixture layer (GML) and a new attention pooling layer proposed in the paper. After obtaining the graph representations, we adopt a hierGCN to classify the graphs. Specifically, we design a consensus loss that plays a key role in the training process. Finally, a node clustering is carried out according to the parameters of GML. The main contributions are as follows: 
\begin{enumerate*}
\item \textbf{New problem:} We introduce a new problem called Inverse Graph Identification (IGI), which tries to identify the nodes in graphs based on the labels of graphs, and we take a formal discussion of the variants of IGI to attract more research attention on this problem.
\item \textbf{New solution:} We propose an effective model called GMGCN based on GMM and GCN to solve a particular study of IGI problem. To the best of our knowledge, this is the first work to achieve node clustering in graph structure by integrating GMM into GCN.
\item \textbf{New loss:} We propose a consensus loss function to boost the model training through the principle of "same attraction, opposite repulsion". Experiments validate that this consensus loss greatly improves the model effect.
\end{enumerate*}

We validate the proposed GMGCN on various synthetic datasets based on different problem statements and a real-world dataset. The results demonstrate that the proposed method is suitable to solve the IGI task.

\section{Related Works}

\textbf{Graph Identification}. Recently, there is an increasing interest in the graph learning domain. Among all the existing works, GCN is one of the most effective convolution models. A typical GCN model is the message passing neural network (MPNN) proposed by Gilmer \emph{te al}.~\cite{gilmer2017neural} which re-generalize several neural network and graph convolutional network approaches as a general "message-passing" architecture. Many kinds of GCN~\cite{bruna2014spectral,defferrard2016convolutional, kipf2017semi} actually deliver different message propagation functions for GCN. 
Among them, Graph Attention Networks (GAT)~\cite{velivckovic2017graph} first leverages learnable self-attentional layers to aggregate weighted neighbors' information. All of these methods obtain the appropriate node representation for node identification tasks. But to cope with the GI problem, a compact representation on graph level should be utilized. Therefore, pooling strategies are proposed to integrate information over the node representations~\cite{wu2020comprehensive}, such as  max/min pooling~\cite{defferrard2016convolutional}, SortPooling~\cite{zhang2018end}, and so on. In addition, Lin \emph{et al}.~\cite{lin2017structured} propose an learnable attention pooling for weighted averaging of node representations. Although several studies have researched the graph identification problem, current researches are still lack of studies about {\em Inverse Graph Identification}.

\textbf{Graph Clustering.} Graph clustering, a fundamental data analysis task that aims to group similar nodes into the same category, has similar goals as IGI problem. Many real-world applications are cast as graph clustering~\cite{shi2000normalized,white2005spectral,hastings2006community,kim2006customer}. The major strategy of graph clustering is to perform simple clustering algorithms such as $K$-means~\cite{jain2010data} or GMM~\cite{mclachlan1988mixture} on the features embedded from graph embedding~\cite{wang2017community}. With greatly successful achievements of deep learning, more graph clustering studies have resorted to deep learning to learn embedding that capturing both node features and structural relationships~\cite{wu2020comprehensive}. Researchers employ the stacked sparse autoencoder~\cite{tian2014learning}, the variational autoencoder~\cite{kipf2016variational}, or the combination both autoencoder and GCN~\cite{bo2020structural} to obtain graph representations for clustering. Nevertheless, these graph clustering methods do not perform well on IGI due to the lack of attention to graph labels.

\textbf{Multiple-Instance Learning.} Another task that has a similar definition to IGI is Multiple-Instance Learning (MIL). MIL is a variant of inductive machine learning, where each learning example consists of a bag of instances instead of a single feature vector~\cite{foulds2010review}. When obtaining local instances annotations is costly or not possible, but global labels for bags are available, MIL is utilized to train classifiers using weakly labeled data. It has received a considerable amount of attention due to both its theoretical interests and its applicability to real-world problems~\cite{dietterich1997solving,andrews2002multiple,cheplygina2019not,pinheiro2015image}. For example, Wu \emph{et al}.~\cite{wu2015deep} propose deep MIL which adopts max pooling to find positive instances for image classification and annotation. 
Although these MIL methods learn a classifier from the labels of bags to achieve the instance classification, these MIL methods ignore the structural information among instances so that they are not suitable for IGI.

\textbf{Gaussian Mixture Model.} In addition, as a core part of the proposed GMGCN model, GMM is a parametric probability density function for a weighted sum of Gaussian component densities~\cite{mclachlan1988mixture}. It is commonly used to find underlying clusters in data samples~\cite{bishop2006pattern}. Generally, the GMM parameters are estimated using the iterative Expectation-Maximization (EM)~\cite{moon1996expectation} algorithm from training data. In this paper, we integrate the GMM into GCN to establish a solution for IGI and update the GMM parameters using stochastic gradient descent. 



\section{Notations and Problem Statement}
\textbf{Notations.} 
We denote a set of graph instances as $G=\{(\mathcal{G}^{(1)},y^{(1)}),\dots,(\mathcal{G}^{(N)}, y^{(N)})\}$ with $N$ graphs, where $\mathcal{G}^{(n)}$ refers to the $n$-th graph instance and $y^{(n)}\in\{0,1,\dots,C-1\}$ is the corresponding graph label of $C$ different categories. 
We denote by $\mathcal{G}^{(n)}=(\mathbb{V}^{(n)},\mathcal{E}^{(n)})$ the $n$-th graph instance of size $M_n$ with nodes $v^{(n)}_i\in \mathbb{V}$ and edges $(v^{(n)}_i,v^{(n)}_j)\in \mathcal{E}^{(n)}$, by $\mathrm{\mathbf{X}}^{(n)}=\{x^{(n)}_1,\dots,x^{(n)}_{M_n}\}\in \mathbb{R}^{M_n\times d}$ the feature matrix of nodes $\mathbb{V}^{(n)}$, and by $\mathrm{\mathbf{A}}^{(n)}\in \{0,1\}^{M_n\times M_n}$ the adjacency matrix which associate edge $(v^{(n)}_i, v^{(n)}_j)$ with $A^{(n)}_{i,j}$. We denote by $\{z^{(n)}_1,\dots,z^{(n)}_{M_n}\}\in \{0,1,\dots,C-1\}$ the potential labels of nodes $\mathbb{V}^{(n)}$, which are invisible to the model training. In addition, we denote by $\mathrm{\mathbf{A}}^{hier}\in \{0,1\}^{N\times N}$ the adjacency matrix of the links among graphs if the graphs in $G$ contain inter-connections. Then, $\mathrm{\mathbf{A}}^{hier}=\mathrm{\mathbf{I}}$ if no relationships among graphs.



\textbf{Problem Statement.} 
The {\em Inverse Graph Identification} problem is defined as: Given a set of graph-label pairs $G$ and node features $\mathrm{\mathbf{X}}$, how to infer each label of the $i$-th nodes in the $n$-th graph, $z^{(n)}_{i}$, where $n\in\{1,\dots,N\}$ and $i \in \{1,\dots,M_n\}$? Under the definition of IGI, there are some different cases that could be extended from the problem as new tasks.
\begin{case}[Infer Original Nodes]
\label{case:orig_nodes}
Given the whole $G$ and $\mathrm{\mathbf{X}}$ as training data, how to infer all the node labels $z^{(n)}_{i}$ in the $\mathcal{G}^{(n)}$s?
\end{case}
\begin{case}[Infer New Nodes]
\label{case:new_nodes}
Given the portion of nodes $\mathbb{V}^{(n)}_{train} = \{v^{(n)}_1,\dots,v^{(n)}_{M_n-s}\}$ in each $\mathcal{G}^{(n)}$ as training data, how to infer the labels of the rest of $s$ nodes $\{v^{(n)}_{M_n-s+1},\dots,v^{(n)}_{M_n}\}$ in the $\mathcal{G}^{(n)}$s?
\end{case}
\begin{case}[Infer New Graphs]
\label{case:new_graph}
Given the portion of graphs $G_{train} = \{(\mathcal{G}^{(1)},y^{(1)}),\dots,(\mathcal{G}^{(N-S)}, y^{(N-S)})\}$ as training data, how to infer the node labels $z^{(n)}_{i},n\in\{N-S+1,\dots,N\}$ in the rest of $S$ graphs, $\{\mathcal{G}^{(N-S+1)},\dots,\mathcal{G}^{(N)}\}$?
\end{case}
\begin{case}[With/Without Hierarchical Graph]
\label{case:hier}
Given the inter-connections among graphs as $\mathrm{\mathbf{A}}_{hier}$, how to infer the node labels. And how to infer the node labels if $\mathrm{\mathbf{A}}_{hier}=\mathrm{\mathbf{I}}$ (i.e. no inter-connections)?
\end{case}

There are more different cases under IGI problem. For example, how to infer a node score instead of a node label? We will discuss more cases in Appendix A. In this paper, we focus on the IGI problem that aims to infer the labels of the nodes. In the Experiments section, we will examine the proposed method and the baselines under all the above cases.

\begin{figure}[t]
  \centering
  \includegraphics[width=0.9\columnwidth]{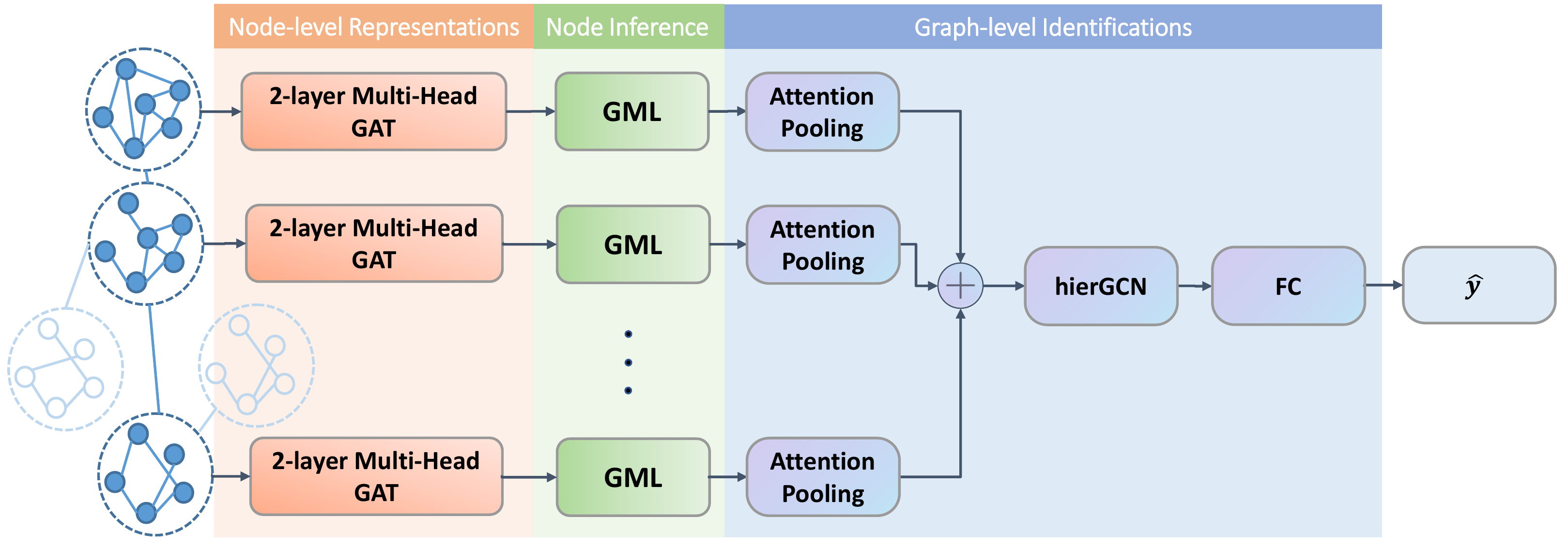}
  \vspace{-2ex}
  \caption{\footnotesize The overall framework of the proposed GMGCN consists of three processes: node-level representations, node inference, and graph-level identifications.}
  \vspace{-4ex}
  \label{fig:framework}
\end{figure}

\section{Proposed Method}
In this section, we first introduce the preliminaries related to the proposed model, then we discuss two key components of GMGCN: GML and consensus loss. 

\subsection{Preliminaries}
\textbf{Graph Attention Networks.} Graph Attention Networks (GAT)~\cite{velivckovic2017graph} has been widely adopted in the field of graph convolution due to the learnable attentions to neighbors during the node update process. A multi-head GAT Convolutional layer (GATConv) is formulated as below:
\begin{equation}
  \vec{h}_i = \mathop{\parallel}^K_{k=1}\sigma(\alpha^k_{i,i}\bm{\Theta}\vec{x}_i + \mathop{\sum}_{j\in N(i)}\alpha^k_{i,j}\bm{\Theta}\vec{x}_j),
\end{equation}
where $\vec{h}_i$ and $\vec{x}_i$ refer to the hidden feature representation and the original feature vector of node $i$ in the graph, respectively. $\parallel$ refers to concatenation operation, and $\sigma( \cdot )$ refers to the nonlinear activation function. $\bm{\Theta}$ is the weight matrix and $N(i)$ is the set of neighbors of node $i$. $\alpha^k_{i,j}$ is the attention coefficient computed by the $k$-th attention mechanism as follows:
\begin{equation}
  \alpha_{i,j}=\frac{\exp({\rm LeakyReLU}({\vec{\rm a}^{\top}}[\bm{\Theta}\vec{x}_i\parallel\bm{\Theta}\vec{x}_j]))}{\mathop{\sum}_{l\in N(i)\cup\{i\}}\exp({\rm LeakyReLU}(\vec{\rm a}^{\top}[\bm{\Theta}\vec{x}_i\parallel\bm{\Theta}\vec{x}_l]))},
\end{equation}
where attention mechanism is a single-layer feedforward neural network, parametrized by a weight vector ${\rm a}\in \mathbb{R}^{2d_2}$, and applying the LeakyReLU nonlinearity. 
In this paper, a two-layer GAT is formulated as follows:
\begin{equation}\label{eq:gat}
  \textbf{H}={\rm GAT}(\textbf{X})=\bm{\alpha}(\mathop{\parallel}^K_{k=1}\sigma(\bm{\alpha}^k\textbf{X}\bm{\Theta}_{a1}))\bm{\Theta}_{a2}.
\end{equation}

\textbf{Graph Attention Pooling.} Lin \emph{et al}. propose the graph attention pooling in \cite{lin2017structured}, which employs a self-attentive mechanism to learn the node importance and then transform a variable number of nodes into a fixed-length graph representation: 
\begin{equation}\label{nodeimpor}
  \textbf{E}={\rm Attn}(\textbf{H})={\rm softmax}(\bm{\Theta}_{s2}{\rm tanh}(\bm{\Theta}_{s1}\textbf{H}^\top))\textbf{H}.
\end{equation}
The node attention scores are calculated by a 2-layer MLP without bias.
$\bm{\Theta}_{s2}$ is employed to infer the importance of each node. ${\rm softmax}$ function is adopted to normalize the importance of each node in the graph. Finally, the unified graph representation of the graph is obtained by multiplying the node attention scores with $\textbf{H}$.

\textbf{Hierarchical GCN.} 
We employ a hierarchical GCN~\cite{li2019semi} to update the feature representations of the graphs in the hierarchical graph as Eq.~\ref{GCNConv}
\begin{equation}\label{GCNConv}
  \textbf{H}'={\mathrm{hierGCN}}(\textbf{H})=\mathrm{Concat}\left(\textbf{H},\sigma(\textbf{A}^{hier}\textbf{H}\bm{\Theta})\right),
\end{equation}
where the hierarchical feature matrix $\textbf{H}$ consists of the feature representations of graphs.
Then we concatenate the original hierarchical feature matrix with the updated feature matrix as a self-loop.

\subsection{Overall Framework}
Fig.~\ref{fig:framework} gives an overall framework of the proposed GMGCN method to infer the categories of the nodes in the graphs. It consists of three processes: node-level representations, node inference, and graph-level identifications.

\textbf{Gaussian Mixture Layer (GML).} As the key component of the GMGCN, GML plays a crucial role in the process of node inference. It is described as $\vec{h}' = \frac{1}{C}\mathop{\sum}^C_{c=1}w_c(\bm{\Theta}_{GML}\vec{h})$, where $\bm{\Theta}_{GML}$ refers to the weight matrix and $w_c(\cdot)$ represents the Gaussian weight function as follows:
\begin{equation}
  w_c(\vec{h})=\exp(-\frac{1}{2}(\vec{h}-\vec{\mu}_c)^\top\bm{\Sigma}^{-1}_c(\vec{h}-\vec{\mu}_c)),
\end{equation}
where $\Sigma_c$ and $\vec{\mu}_c$ are learnable covariance matrix and mean vector of the Gaussian weight function $w_c(\cdot)$, respectively. 

Based on the idea of Gaussian Mixture Model (GMM), GML has the ability to effectively distinguish different types of nodes according to the input features. However, it could not resolve two major challenges of IGI problem: the node relations and the graph identification that affects the final node inference. Therefore, as shown in Fig.~\ref{fig:framework}, we employ the node representations from the 2-layer GAT as inputs of GML. The node representations provide local structural information of the nodes to GML. Meanwhile, the outputs $\vec{h}'$ of GML are aggregated by attention pooling and further fed into a hierarchical GCN for graph identifications. The graph identifications adjust the hidden space of GML via a back propagation process of hierGCN. Note that, one can easily replace hierGCN by multilayer perceptron (MLP) to formulate Case\ref{case:hier} of IGI. Both node representations and graph identifications address two major challenges of IGI, respectively.

\begin{wrapfigure}[13]{r}{0.4\textwidth}
    \centering
    \includegraphics[width=0.4\textwidth]{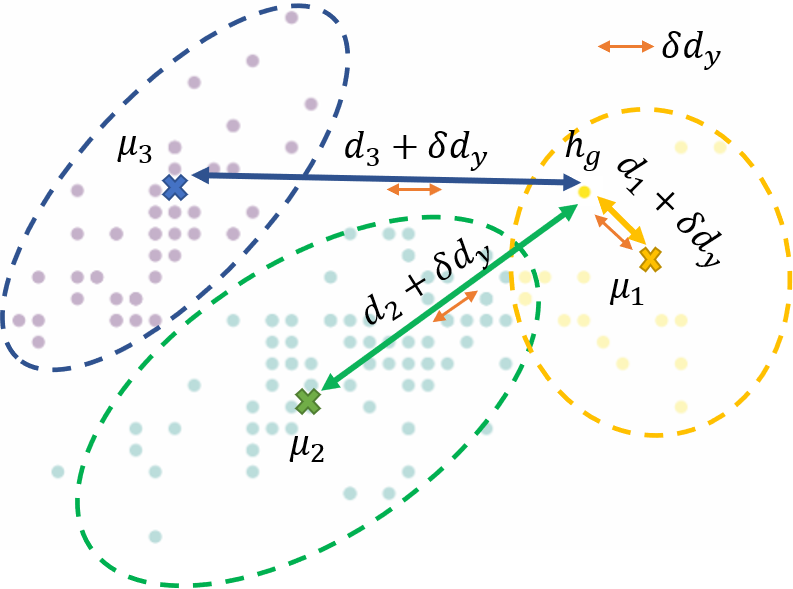}
  \caption{\footnotesize Calculation of the enhanced distance matrix $(d_{nc})_{enhance}$.}
  \label{fig:conloss}
\end{wrapfigure}
\textbf{GMGCN vs. AutoEncoder.} Actually, the GMGCN model can be re-thought as an encoder-decoder process with a pre-specified hidden space, where the 2-layer GAT is considered as an encoder, the hierGCN is considered as a decoder to graph labels instead of nodes themselves, and GML is the hidden space. When involving GML into the proposed model, it naturally involves an assumption that the nodes of each category are potentially distributed as a Gaussian distribution in the hidden space. The nodes of multiple categories are mixed out to form a Gaussian mixture distribution. Consequently, this also implies the limitation of the proposed model that the proposed model is only able to resolve the IGI problem for categorizing nodes. 

\textbf{Consensus Loss.} 
The core idea of consensus loss is to make the graph representations with the same labels closer, and those with different labels farther away in the hidden space. Therefore, before computing the consensus loss, we first compute the similarity matrix $\mathrm{\mathbf{S}}\in \mathbb{R}^{N\times C}$ of the graph representations, which is obtained by the following two steps:

First, we compute the distance between the graph representation $\vec{h}^n_{g}$ via graph attention pooling layer and the mean vector $\vec{\mu}_c$ of $c$-th Gaussian weight function as follows:
\begin{equation}\label{dis}
  d_{nc}=\parallel\vec{h}^n_{g}-\vec{\mu}_{c}\parallel_2.
\end{equation}
Second, we enhance the distance of the graph representations to the true mean vector on $d_{nc}$ by:
\begin{equation}\label{norm}
  (d_{nc})_{enhance}=(d_{nc}+\delta d_{ny^{(n)}}),
\end{equation}
where $\delta$ is a discount hyperparameter, $y^{(n)}$ is the true label of graph $\mathcal{G}^{(n)}$, and $d_{ny^{(n)}}$ represents the distance between the graph representation $\vec{h}^n_{g}$ and the mean vector of $y^{(n)}$-th Gaussian weight function $\vec{\mu}_{y^{(n)}}$. 
Then the similarity matrix $\textbf{S}$ is formulated as $\textbf{S}={\rm softmax}(-\mathrm{\mathbf{D}}_{enhance})$, where $\mathrm{\mathbf{D}}_{enhance} \in \mathbb{R}^{N\times C}$ consists of $(d_{nc})_{enhance}$ and is calculated as Fig.~\ref{fig:conloss}.

Finally, the consensus loss, $L_{con}$, is formulated as a cross-entropy loss between $s_{nc}$ and $y^{(n)}$. 
In this way, the consensus loss makes the graph representations of each graph closer to the mean vector of the Gaussian function corresponding to its graph label, so as to realize that the graphs with the same labels are closer and the graphs with different labels are farther. Meanwhile, the graph attention pooling allows the node representations with the potentially opposite label to their graph labels to leave away from the graph representations by paying less attention.

\textbf{Overall Loss.} Besides the consensus loss that distinguishes nodes in the hidden space, the GMGCN model also includes the graph classification task. We establish a cross-entropy loss $L_{cls}$ between the predictions and the ground truth labels over all graphs. Then, the overall loss function of our GMGCN is defined as linear interpolation of $L_{cls}$ and $L_{con}$ as $L=\alpha L_{cls}+(1-\alpha) L_{con}$, where $\alpha \in [0,1]$ is the trade-off coefficient between classification loss $L_{cls}$ and consensus loss $L_{con}$.
\subsection{Inference of Nodes}
We infer the categories of the nodes in the graph based on the cosine similarity between the node representations, $\vec{h}$, obtained from GML and the mean vector, $\vec{\mu}_c$, of each Gaussian weight function. The category $\hat{z}$ of each node is determined as follows:
\begin{equation}
  \hat{z}={\rm arg}\mathop{\rm max}_c cosine(\vec{h},\vec{\mu}_c).
\end{equation}

\begin{table}[tbp]
  \scriptsize
  \renewcommand{\arraystretch}{1.1}
  \centering
  \caption{Comparison of different types of methods on Synthetic dataset (mean(std)).}
    \begin{tabular}{c|cc|cc|cc}
    \toprule
    \multirow{3}[6]{*}{methods} & \multicolumn{6}{c}{ single graph} \\
\cmidrule{2-7}          & \multicolumn{2}{c|}{orignal nodes} & \multicolumn{2}{c|}{new nodes} & \multicolumn{2}{c}{new graphs} \\
\cmidrule{2-7}          & NMI   & ARI   & NMI   & ARI   & NMI   & ARI \\
    \midrule
    Feature Clustering & 49.46(1.89) & 50.78(2.48) & 50.11(2.07) & 49.33(2.56) & 47.48(2.48) & 48.03(3.32) \\
    Graph Clustering  & 80.65(8.18) & 85.51(7.96) & 85.85(8.47) & 87.72(8.52) & 78.72(10.47) & 84.25(10.33) \\
    MIL & 76.47(0.00) & 81.39(0.00) & 75.42(0.00) & 80.76(0.00) & 75.77(0.00) & 80.63(0.00) \\ 
    \midrule
    ATTGCN & 14.25(9.76) & 12.45(11.64) & 19.69(9.54) & 13.44(11.38) & 13.25(9.60) & 12.22(11.10) \\
    GMGCN-noncon & 49.58(37.15) & 56.57(31.17) & 60.46(26.25) & 57.72(31.87) & 54.91(25.69) & 56.88(30.34) \\
    GMGCN & \textbf{93.28(0.77)} & \textbf{96.29(0.43)} & \textbf{96.70(0.43)} & \textbf{97.55(0.34)} & \textbf{93.49(1.12)} & \textbf{96.46(0.62)} \\
    \midrule
    \midrule
    \multirow{3}[6]{*}{methods} & \multicolumn{6}{c}{hierarchical graph} \\
\cmidrule{2-7}          & \multicolumn{2}{c|}{orignal nodes} & \multicolumn{2}{c|}{new nodes} & \multicolumn{2}{c}{new graphs} \\
\cmidrule{2-7}          & NMI   & ARI   & NMI   & ARI   & NMI   & ARI \\
    \midrule
    ATTGCN & 19.09(12.34) & 17.48(13.52) & 24.64(12.34) & 19.09(13.23) & 20.54(11.92) & 20.00(13.37) \\
    GMGCN-noncon & 46.45(17.61) & 37.44(38.26) & 48.08(28.28) & 37.66(38.33) & 49.17(25.88) & 37.26(38.14) \\
    GMGCN & \textbf{92.83(1.22)} & \textbf{96.02(0.71)} & \textbf{96.32(0.77)} & \textbf{97.20(0.62)} & \textbf{92.54(1.71)} & \textbf{95.95(0.88)} \\
    \bottomrule
    \end{tabular}%
    \vspace{-3ex}
  \label{tab:newnodegraph}%
\end{table}%

\section{Experiments}
In this section, we first conduct the experiments on synthetic data to validate the effectiveness of our proposed model under different problem cases. Then, we construct a real-world dataset based on the public rumor dataset called PHEME\footnote{https://figshare.com/articles/PHEME\_dataset\_of\_rumours\_and\_non-rumours/4010619}~\cite{zubiaga2016learning}, the proposed method and the comparisons are evaluated on the constructed dataset.

\subsection{Baselines}
Since the IGI problem is a new challenge without any baseline work yet, we compare the proposed GMGCN with three types of most similar methods: clustering methods without structure information~\cite{jain2010data,mclachlan1988mixture,ester1996density,hinton2006reducing}, graph clustering methods~\cite{kipf2016variational,bo2020structural} and MIL~\cite{ray2005supervised} methods. We only report the best results of each type of method in the sequel due to the page limitations. More detailed results for all methods are available in Appendix B. 

\textbf{Proposed Method.} We also examine the various settings of the proposed method as:
\begin{itemize*}
      \item \textbf{ATTGCN}: GMGCN without GML, we replace cosine similarity by 2-head attention mechanisms in the attention pooling layer to distinguish the categories of the nodes in the inference.
      \item \textbf{GMGCN}: The proposed method. 
      \item \textbf{GMGCN}-noncon: GMGCN without the proposed consensus loss for training.
\end{itemize*} 

We apply Normalized Mutual Information (NMI)~\cite{strehl2002cluster} and Adjusted Rand Index (ARI)~\cite{hubert1985comparing} to evaluate the clustering results in this paper. 
The dimension of our GMGCN is set to \emph{d}-128-64-16-64-\emph{c}, where \emph{d} is the input dimension and \emph{c} is the number of graph categories in the dataset. We adopt 2 independent attention mechanisms for the first GATConv layer and use 1 attention mechanism for the second GATConv layer. The parameters are updated using stochastic gradient descent via Adam algorithm \cite{kingma2014adam}. The discount hyperparameter $\delta$=0.5. The training process is iterated upon 800 epochs for GMGCN. We run all methods 10 times to avoid extremes and report the average results.

\begin{figure}[t]
\centering
\subfigure[Original features]{
\label{orgn}
\begin{minipage}[t]{0.22\linewidth}
\centering
\includegraphics[width=1\columnwidth]{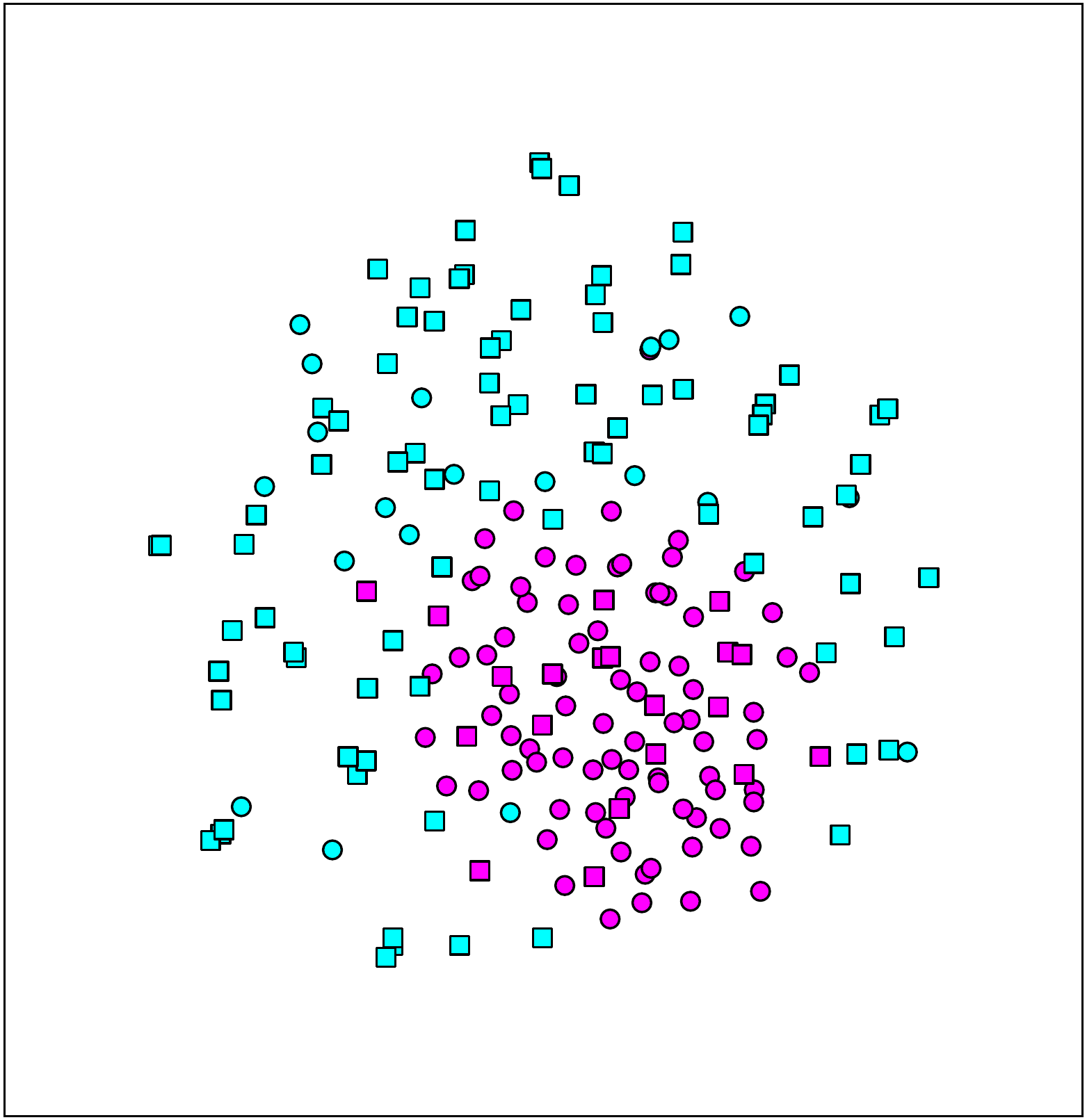}
\end{minipage}
}
\subfigure[Feature Clustering]{
\label{AE}
\begin{minipage}[t]{0.22\linewidth}
\centering
\includegraphics[width=1\columnwidth]{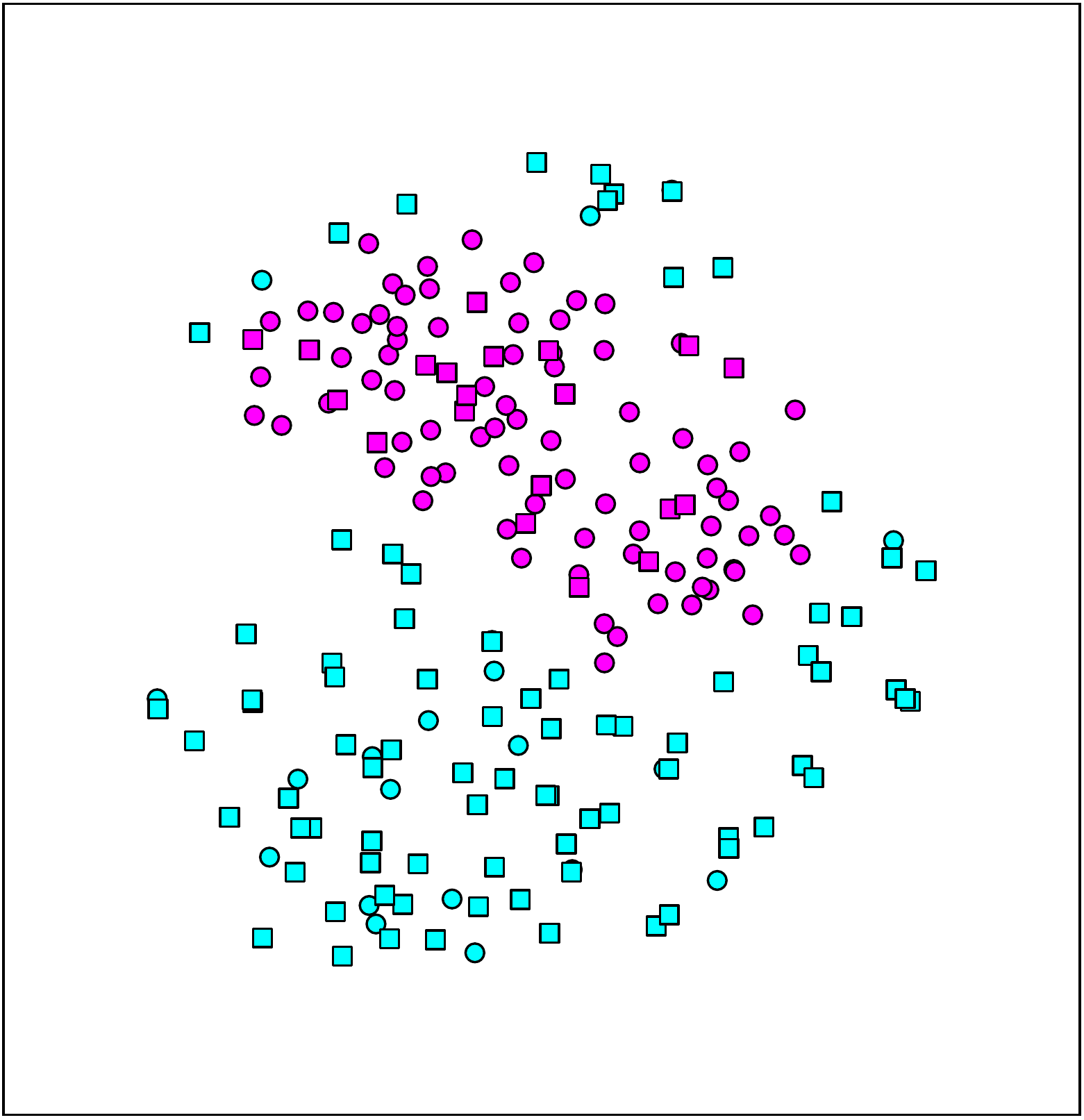}
\end{minipage}
}
\subfigure[Graph Clustering]{
\label{GAE}
\begin{minipage}[t]{0.22\linewidth}
\centering
\includegraphics[width=1\columnwidth]{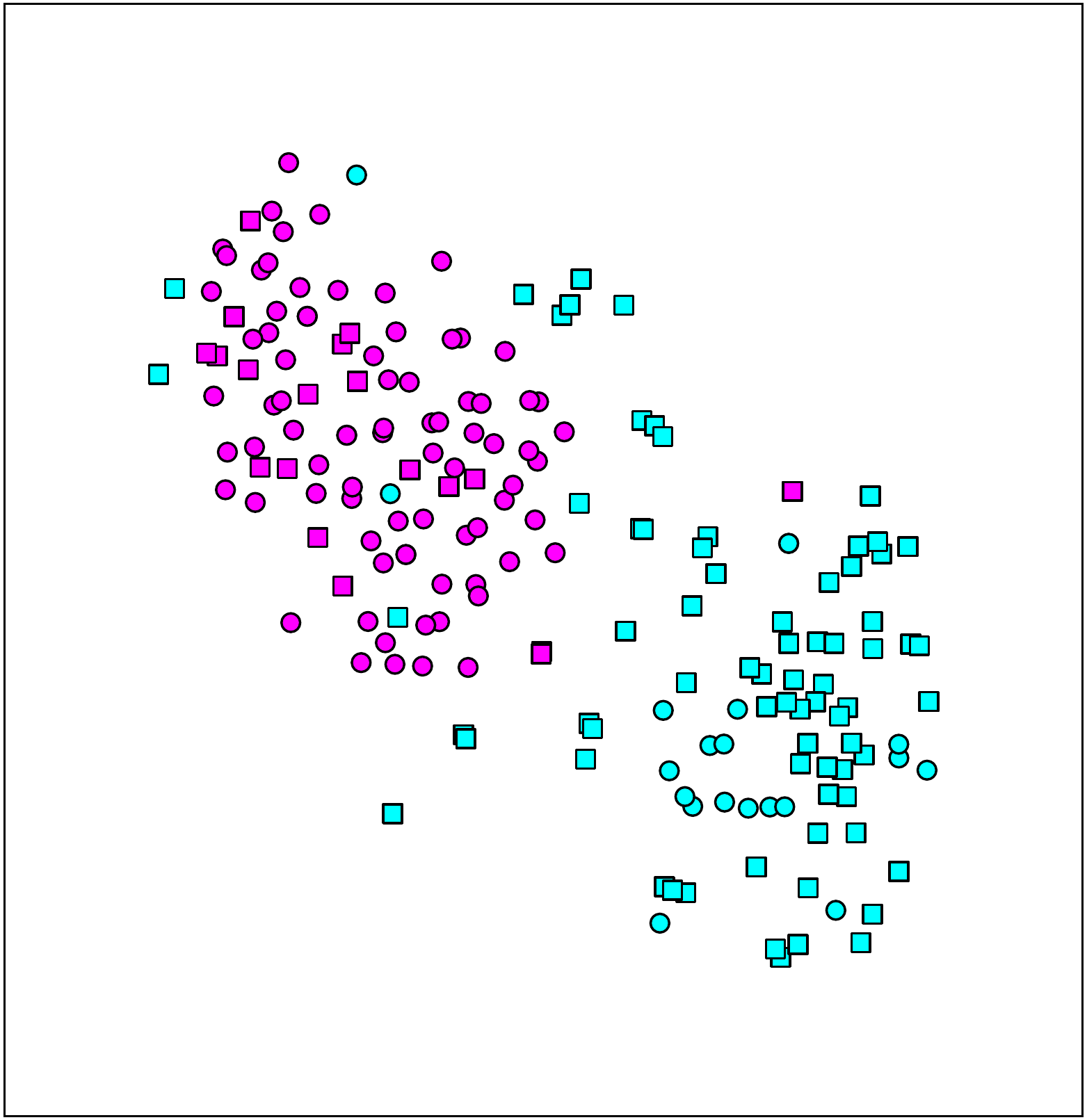}
\end{minipage}
}
\subfigure[GMGCN]{
\label{GMGCN}
\begin{minipage}[t]{0.22\linewidth}
\centering
\includegraphics[width=1\columnwidth]{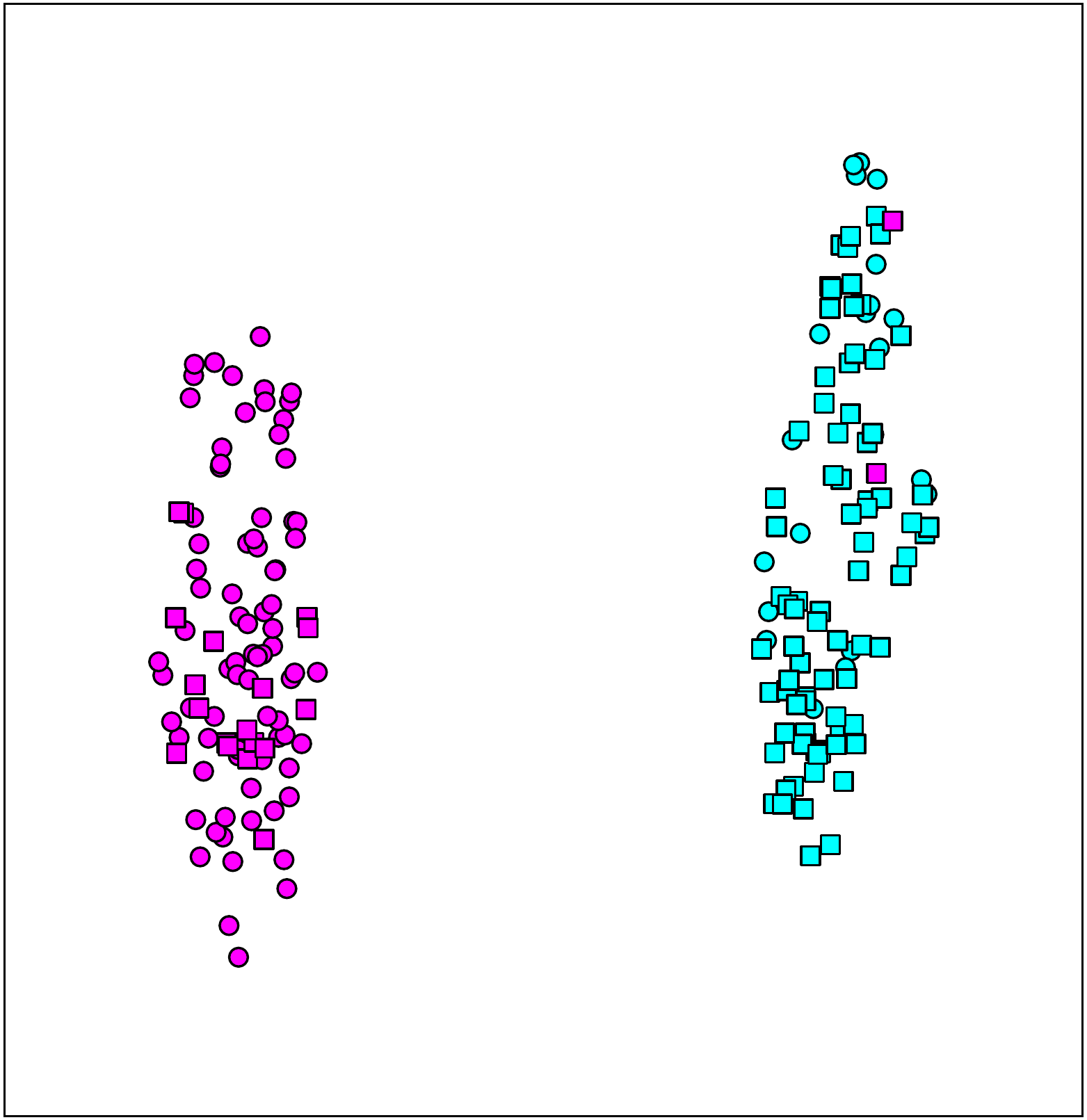}
\end{minipage}
}
\vspace{-2ex}
\footnotesize
\caption{2D visualization of the original node features, the node representations learned by feature clustering, graph feature clustering, and GMGCN from two randomly sampled graphs. The different shapes of the symbols represent different graphs and the different colors represent different node categories.}
\vspace{-4ex}
\label{vis}
\end{figure}

\subsection{GMGCN on Synthetic Data}
\textbf{Synthetic Data Generation.} 
We generate a synthetic hierarchical graph to examine the proposed IGI problem with various cases. First, we generate two synthetic user groups, representing normal and abnormal users, respectively. The relationship between normal users is generated by the Barabási–Albert graph model~\cite{barabasi1999emergence}, while the relationship between abnormal users is generated by the random graph algorithm. We use Gaussian distribution to randomly generate user features, the mean and standard deviation of these two types of users are randomly sampled from [-5, 5] and [1, 10], respectively. Second, we randomly connect users between two groups to simulate the relationship between two different types of users. 
Third, we sample $20\%$ of abnormal (normal) users into the normal (abnormal) graph so that a graph structure consists of two different types of users. We sample 100 users per graph, and generate 50 normal graphs and 50 abnormal graphs. The label of each graph is determined by the major labels of the users in the graph. Finally, we construct the connections between two graphs if they have more than one common user.


\textbf{Case~\ref{case:orig_nodes}: Transductive learning.} All graphs are available in the training. Table~\ref{tab:newnodegraph} shows that the proposed GMGCN method outperforms all the baselines. In particular, GMGCN improves NMI and ARI by more than 10\% compared with all best-performing baseline results. This demonstrates a great improvement of the node identifications by the graph labels.

\textbf{Case~\ref{case:new_nodes} \& Case~\ref{case:new_graph}: Inductive learning.} Portions of graphs or nodes are available in the training. For the testing of new nodes and new graphs, we sample $20\%$ of the nodes from each graph and $20\%$ of the graphs as testing, respectively. GMGCN also outperforms the other baselines as shown in Table~\ref{tab:newnodegraph}. Therefore, GMGCN works for both transductive and inductive settings.

\textbf{Case~\ref{case:hier}: With hierarchical graph.} As shown in Table ~\ref{tab:newnodegraph}, the hierarchical graph structure promotes the performance of ATTGCN but has no effect on GMGCN. This indicates that GMGCN focuses more on node-level updates than graph-level updates to obtain better node clustering results.

\textbf{Attention vs. GML.} We compare the ATTGCN with GMGCN to verify the importance of GML. The results show that GML greatly promotes GMGCN to identify nodes, while the attention mechanism fails.

\textbf{Consensus loss.} To analyze the effect of the proposed consensus loss, we compare the performance of GMGCN and GMGCN-noncon. The results in Table~\ref{tab:newnodegraph} 
indicates that consensus loss enables GMGCN to identify node categories well. 

Consequently, the out-performances of GMGCN for all cases imply that the proposed method is a right solution for IGI problem.


\textbf{Visualization.} 
We also visualize the node representations of two graphs randomly sampled from the Synthetic dataset in a 2D space by t-SNE~\cite{van2014accelerating}. Figs~\ref{orgn}-\ref{GMGCN} show the original node features, as well as the node representations learned from the feature clustering, graph clustering, and GMGCN, respectively. Different shapes in the figures represent the nodes from different graphs and different colors refer to different node categories. As shown in Fig.~\ref{GMGCN}, GMGCN clearly distinguishes these two node categories but the others fail. This demonstrates that the node representations learned by GMGCN are reasonable for IGI.

\begin{table}[tbp]\scriptsize
  \centering
  \caption{Comparison of different types of methods on PHEME datasets (mean(std)).}
    \begin{tabular}{c|ccccc}
    \toprule
    \multicolumn{1}{c|}{} & Charlie Hebdo & Ferguson & Germanwings Crash  & \multicolumn{1}{c}{Ottawa Shooting} & Sydney Siege \\
    \midrule
    Method & NMI   & NMI   & NMI   & NMI & NMI \\
    \midrule
    Feature Clustering & 25.46(0.00) & 24.56(0.00) & 48.84(0.00) & \multicolumn{1}{c}{\textbf{34.56(0.00)}} & 19.45(0.00) \\
    Graph Clustering   & 3.26(2.26) & 1.08(0.71) & 0.66(0.50) & \multicolumn{1}{c}{4.49(3.07)} & 4.93(1.92) \\
    MIL & 5.69(0.00) &  4.08(0.00)   & 0.61(0.00)  &  0.60(0.00)   & 19.68(0.00) \\
    GMGCN & \textbf{47.51(3.27)} & \textbf{48.35(4.08)} & \textbf{48.85(2.14)} & 32.58(3.63) & \textbf{41.00(3.93)} \\
    \midrule
    \midrule
    Method & ARI   & ARI   & ARI   & ARI & ARI \\
    \midrule
    Feature Clustering & 23.46(0.00) & 23.23(0.00) & 47.29(0.00) & 32.08(0.00) & 16.45(0.00) \\
    Graph Clustering    & 6.56(6.30) & 4.04(2.24) & 3.71(1.93) & 10.66(7.68) & 11.78(5.57) \\
    MIL & 9.38(0.00) &  13.93(0.00)   & 1.10(0.00)  &  2.09(0.00)   & \textbf{39.51(0.00)} \\
    GMGCN & \textbf{52.38(3.15)} & \textbf{55.26(3.21)} & \textbf{54.95(1.02)} & \textbf{37.51(2.23)} & 37.79(2.18) \\
    \bottomrule
    \end{tabular}%
    \vspace{-4ex}
  \label{tab:clurumor}%
\end{table}%

\subsection{GMGCN on PHEME Data}
\textbf{Data Description.} 
PHEME consists of five sets of real-life tweets, where each set is related to a piece of breaking news. 
Every breaking news includes a lot of rumor and non-rumor topics. A detailed data description is presented in~\cite{zubiaga2016learning}. Based on this dataset, we construct the hierarchical graph structure by assuming each topic as a graph, where links between users are formed according to their follows/retweets. We connect two graphs if they have more than one common user. We label the users appearing in more than \emph{M} rumor topics as abnormal users, and the rest as normal users. Therefore, this dataset perfectly falls into the statement of the IGI problem. When $M = 2, 3, 4$, the proposed method all achieves the best results, the detailed results are shown in Appendix B. In this section, we only show the results when $M=4$.

\textbf{Clustering Results.} 
As shown in Table~\ref{tab:clurumor}, GMGCN perfectly resolves this IGI problem with an average $13\%$ improvement on NMI and a $15\%$ improvement on ARI on these five PHEME datasets compared with the best-performing baseline methods. This significant improvement implies that there exist certain applications in the real world that fulfill the statement of IGI problem. In contrast, the other methods perform poorly because IGI is a totally new problem that differs from feature clustering, graph clustering, or MIL. Due to the improper assumptions, they all fail to reveal the truth.

\section{Discussion and Future Work}
In this work, we explore an interesting and novel problem: Can we identify nodes given the identifications of graphs? In one word, can we {\em Invert the Graph Identification}? The proposed IGI problem has totally different notions and statements from any of the node clustering, graph clustering, or multiple instance learning tasks. To address this new issue, we propose GMGCN with the integration of GMM and GNNs to resolve certain cases of IGI problem. Experimental results on Synthetic data and real-world data reveal the need for formulating IGI and the advantage of the proposed GMGCN over other related baselines.
Nevertheless, our study is just an initial step towards the IGI problem while a variety of extensions are still potential. For example, can we infer the labels of edges or sub-graphs instead of nodes? Can we detect anomaly nodes other than conducting node clustering? We hope our study will open up a new vein of graph learning and encourage more specifications, solutions, and developments for IGI.








\section*{Broader Impact}
Nowadays, many studies have been developed on Graph Identification (GI), such as graph classification and graph regression. However, the inverse problem of using the information of graphs to infer the information of nodes or even edges and sub-graphs has never been raised. In this paper, we present a new challenge named as Inverse Graph Identification (IGI). As opposed to GI, IGI mainly explores whether we can identify nodes given the labels of graphs they belong to. We also analyze various tasks belonging to the IGI problem so that more researchers can participate in the research of the IGI problem. By studying the IGI problem, many graph learning problems will be solved in a more appropriate and efficient way. The broader impact of our research can be summarized below:

\textbf{For social media community:} Through the research of social media tasks on IGI problem, different types of information communicators can be identified more accurately. For example, media reporters can use the proposed model to find appropriate interviewees, advertisers can discover the users in social networks for accurate advertising.

\textbf{For biology and chemistry community:} Biologists can divide proteins into groups according to the different interactions of proteins, and then use our method to find proteins with specific functions. Chemists can employ our method to discover functional groups with special properties. 

\textbf{For network security community:} By dividing different people into groups, network security officers can utilize our method to discover suspicious users on the LAN, and can also employ our method to find suspicious operations or applications by grouping different operations or software together.

\bibliographystyle{unsrt}
\bibliography{myreference}
\newpage
\appendix
\section{Supplementary Materials}
\subsection{Problem Statement}
In this section, we will introduce some other cases under IGI problem. Most of them are still open problems that are waiting for further researches.

\textbf{Notations.} We denote a set of graph instances as $G=\{(\mathcal{G}^{(1)},y^{(1)}),\dots,(\mathcal{G}^{(N)}, y^{(N)})\}$ with $N$ graphs, where $\mathcal{G}^{(n)}$ refers to the $n$-th graph instance and $y^{(n)}\in\{0,1,\dots,C-1\}$ is the corresponding graph label of $C$ different categories. 
We denote by $\mathcal{G}^{(n)}=(\mathbb{V}^{(n)},\mathcal{E}^{(n)})$ the $n$-th graph instance of size $M_n$ with nodes $v^{(n)}_i\in \mathbb{V}$ and size $T_n$ with edges $(v^{(n)}_i,v^{(n)}_j)\in \mathcal{E}^{(n)}$, by $\mathrm{\mathbf{X}}^{(n)}_V \in \mathbb{R}^{M_n\times d_v}$ the feature matrix of nodes $\mathbb{V}^{(n)}$, by $\mathrm{\mathbf{X}}^{(n)}_E\in \mathbb{R}^{T_n\times d_e}$ the feature matrix of edges $\mathcal{E}^{(n)}$, and by $\mathrm{\mathbf{A}}^{(n)}\in \{0,1\}^{M_n\times M_n}$ the adjacency matrix which associate edge $(v^{(n)}_i, v^{(n)}_j)$ with $A^{(n)}_{i,j}$. We denote by $\{z^{(n)}_1,\dots,z^{(n)}_{M_n}\}\in \{0,1,\dots,C-1\}$ the potential labels of nodes $\mathbb{V}^{(n)}$, by $\{s^{(n)}_1,\dots,s^{(n)}_{M_n}\}\in [0,1]$ the potential scores of nodes $\mathbb{V}^{(n)}$, by $\{e^{(n)}_1,\dots,e^{(n)}_{T_n}\}\in \{0,1,\dots,C-1\}$ the potential labels of edges $\mathcal{E}^{(n)}$, and by $\{o^{(n)}_1,\dots,o^{(n)}_{R_n}\}\in \{0,1,\dots,C-1\}$ the potential labels of sub-graphs, which are invisible to the model training. In addition to the cases mentioned in the Notations and Problem Statement Section, more cases under IGI problem are introduced as below:
\begin{case}[Infer Nodes Based on Semi-supervised Learning]
\label{case:semi}
  Given the whole $G$, $\mathrm{\mathbf{X}}$ and the portion of the node labels $\{z^{(n)}_1,\dots,z^{(n)}_{M_n-s}\}$ as training data, how to infer the labels of the rest of nodes $\{z^{(n)}_{M_n-s+1},\dots,z^{(n)}_{M_n}\}$ in the $\mathcal{G}^{(n)}$s?
  \end{case}
  
It is a very common situation. For example, in the PHEHE data, we may be able to know a portion of users' labels. Then, the IGI problem falls into Case~\ref{case:semi}. But to cope with Case~\ref{case:semi}, the proposed GMGCN needs a designated loss to gain extra information from the labeled nodes in the graphs.


\begin{case}[Node Ranking]
  Given a set of graph-label pairs $G$ and node features $\mathrm{\mathbf{X}}_V$ as training data, how to obtain the node score of the $i$-th node in the $n$-th graph, $s^{(n)}_{i}$, where $n\in\{1,\dots,N\}$ and $i \in \{1,\dots,M_n\}$? 
\end{case}
For example, one application could be to find the influential nodes such as the community leader in the graph.

\begin{case}[Infer Edges]
  Given a set of graph-label pairs $G$, node features $\mathrm{\mathbf{X}}_V$ and edge features $\mathrm{\mathbf{X}}_E$ as training data, how to infer the label of the $i$-th edge in the $n$-th graph, $e^{(n)}_{i}$, where $n\in\{1,\dots,N\}$ and $i \in \{1,\dots,T_n\}$? 
\end{case}
This case could be applied to edge identification, for example, identifying labels of indirect jump instructions (including indirect jump, indirect call, and function return instructions) in a Control Flow Graph (CFG).

\begin{case}[Infer Sub-graphs]
  Given a set of graph-label pairs $G$ and node features $\mathrm{\mathbf{X}}_V$ as training data, how to infer all the sub-graph labels $o^{(n)}_{i}$ in the graphs $\mathcal{G}^{(n)}$s? 
\end{case}
This case refers to a sub-graph identification task, for example, inferring the roles of the functional groups in each molecule given its chemical or physical properties.

\begin{case}[Abnormal Detection]
\label{case:anomaly}
  Given a set of graph-label pairs $G$ and node features $\mathrm{\mathbf{X}}_V$ as training data, where most of $z^{(n)}_i$ belongs to one category as normal nodes but only a few of $z^{(n)}_i$ belongs to the other category as abnormal nodes, 
  how to find out all the abnormal nodes in the graphs $\mathcal{G}^{(n)}$s? 
\end{case}
\begin{table}[tbp]
  \centering
  \caption{Comparison of different methods on SocialGroups dataset (mean(std)). The bold numbers represent the best results.}
    \begin{tabular}{cccccc}
    \toprule
   Method & ACC   & NMI   & ARI   & F1    & Pre@$n$ \\
   \midrule
    K-Means & \textbf{99.13(0.00)} & 0.00(0.00) & 0.00(0.00) & \textbf{49.78(0.00)} & 0.00(0.00) \\
    GMM   & 89.31(29.48) & 0.00(0.00) & 0.00(0.00) & 44.89(14.68) & 0.00(0.00) \\
     DBSCAN  & 0.00(0.00) & 1.11(0.00) & \textbf{2.10(0.00)} & 0.00(0.00) & 1.71(0.00) \\
    AE    & 56.99(0.26) & 5.86(0.06) & 0.46(0.08) & 39.47(0.13) & 2.00(0.13) \\
    \midrule
    GAE   & 56.99(0.26) & 5.86(0.06) & 0.46(0.08) & 39.47(0.13) & 2.00(0.13) \\
    VGAE  & 57.77(1.38) & 5.39(0.13) & 0.18(0.19) & 38.93(0.60) & 2.25(0.29) \\
    SDCN  & 91.58(0.00) & 0.00(0.00) & 0.00(0.00) & 47.66(0.00) & 0.00(0.00) \\
    SIL   & 93.45(0.00) & 0.01(0.00) & 0.25(0.00) & 48.75(0.00) & \textbf{4.59(0.00)} \\
    \midrule
    GMGCN & 76.09(7.55) & \textbf{7.09(2.23)} & 0.02(0.14) & 45.98(1.21) & 0.71(0.80) \\
    \bottomrule
    \end{tabular}%
  \label{tab:abnormal}%
\end{table}%

\textbf{The limitation of GMGCN on Case~\ref{case:anomaly}.} The anomaly detection case under IGI problem is a task that the proposed GMGCN method and related clustering methods fail to cope with. We observe this limitation of the proposed GMGCN method and the baselines on a SocialGroups dataset provided by an anonymous company. The SocialGroups dataset consists of 2039 online social groups, each with no more than 100 users. On average, there are 2.9 abnormal users per graph. Each baseline method are elaborated in the next Section, and we adopt five metrics -- Accuracy (ACC)~\cite{taylor1997introduction}, Normalized Mutual Information (NMI)~\cite{strehl2002cluster}, Adjusted Rand Index (ARI)~\cite{hubert1985comparing}, F1 score (F1)~\cite{van1979information}, and Precision@$n$ (Pre@$n$)~\cite{powers2011evaluation} ($n$ varies with the number of abnormal nodes in each graph)-- to measure the effectiveness of anomaly detection. As shown in Table~\ref{tab:abnormal}, each method performs very poorly, especially on the Pre@n metric. It indicates that for such extremely imbalanced data, the clustering methods, even for GMGCN, are unable to find the abnormal nodes in each graph effectively. Therefore, we need further studies on the anomaly detection case of IGI problem, and put forward a novel model to solve it effectively.

Of course, there are many cases under IGI problem, and more cases are left to other researches to explore.

\subsection{Detailed Experimental Results}
In this section, we present the detailed experimental results corresponding to the Experiment Section. First of all, we introduce the baseline methods mentioned in the Experiment Section. Then, we present the detailed results on synthetic and PHEHE datasets. Our code will be available on GitHub \footnote{https://github.com/IGIproblem/IGIproblem}.
\subsubsection{Baselines}
\textbf{Feature Clustering Methods:}
\begin{itemize}
      \item \textbf{\emph{K}-means}~\cite{jain2010data}: A classical clustering method that aims to partition data points into $K$ clusters in which each data point belongs to the cluster with the nearest cluster centroid.
      \item \textbf{GMM}~\cite{mclachlan1988mixture}: A probabilistic model that assumes all the data points are generated from a mixture of a finite number of Gaussian distributions with unknown parameters.
      \item \textbf{DBSCAN}~\cite{ester1996density}: A density-based algorithm for discovering clusters in large spatial databases with noise.
      \item \textbf{AE}~\cite{hinton2006reducing}: A two-stage deep clustering algorithm. We perform GMM on the representations learned by autoencoder.
\end{itemize} 

\textbf{Graph Clustering Methods:}
\begin{itemize}
      \item \textbf{GAE \& VGAE}~\cite{kipf2016variational}: A structural deep clustering model that combines GCN with the (variational) autoencoder to learn representations. we perform GMM on the representations learned by graph autoencoder.
      \item \textbf{SDCN}~\cite{bo2020structural}: A structural deep clustering network model that integrates the structural information into deep clustering.
\end{itemize} 

\textbf{Multiple Instance Learning:}
\begin{itemize}
      \item \textbf{SIL}~\cite{ray2005supervised}: Single-Instance Learning (SIL) is a MIL approach that assigns each instance the label of its bag, creating a supervised learning problem but mislabeling negative instances in positive bags.
\end{itemize}

We implement $K$-means, GMM and DBSAN with scikit-learn\footnote{https://scikit-learn.org}; AE, GAE, VGAE, SDCN with Pytorch\footnote{https://pytorch.org/}; SIL with a Python implementation created by~\cite{doran2014theoretical}. To be consistent with related works~\cite{bo2020structural}, the dimension of AE is set to \emph{d}-500-500-2000-10, where \emph{d} is the dimension of the input data, and the dimension of GAE and VGAE is set to \emph{d}-256-16. We train AE with 100 epochs, GAE and VGAE with 400 epochs, and SDCN with 200 epochs. 

\subsubsection{Detailed Results}

\subsubsubsection{Synthetic Dataset}
Table~\ref{tab:detailnewnodegraph} shows the detailed experimental results of Table~1 in the Experiment Section. Because of the lack of generalization capability, the DBSCAN method is ineffective in dealing with new nodes and new graphs. 

\begin{table}[htbp]
  \scriptsize
  \renewcommand{\arraystretch}{1.1}
  \centering
  \caption{Comparison of different methods on Synthetic dataset (mean(std)). The bold numbers represent the best results.}
    \begin{tabular}{c|cc|cc|cc}
    \toprule
    \multirow{3}[6]{*}{methods} & \multicolumn{6}{c}{ single graph} \\
\cmidrule{2-7}          & \multicolumn{2}{c|}{orignal nodes} & \multicolumn{2}{c|}{new nodes} & \multicolumn{2}{c}{new graphs} \\
\cmidrule{2-7}          & NMI   & ARI   & NMI   & ARI   & NMI   & ARI \\
    \midrule
    K-Means & 49.46(1.89) & 50.78(2.48) & 50.11(2.07) & 49.33(2.56) & 47.48(2.48) & 48.03(3.32) \\
    GMM   & 50.08(49.92) & 50.08(49.92) & 50.00(50.00) & 49.99(50.01) & 50.00(49.99) & 49.99(50.01) \\
    DBSCAN & 15.42(0.00) & 8.35(0.00) & -     & -     & -     & - \\
    AE    & 5.11(14.33) & 5.36(15.01) & 5.39(16.13) & 5.29(15.94) & 4.73(14.18) & 4.96(14.92) \\
    \midrule
    GAE   & 48.25(37.95) & 48.45(46.03) & 51.43(29.90) & 45.56(41.72) & 29.13(26.07) & 26.89(30.54) \\
    VGAE  & 34.70(37.65) & 31.18(45.44) & 38.71(24.53) & 29.30(34.80) & 16.06(21.99) & 13.13(24.47) \\
    SDCN  & 80.65(8.18) & 85.51(7.96) & 85.85(8.47) & 87.72(8.52) & 78.72(10.47) & 84.25(10.33) \\
    \midrule
    SIL & 76.47(0.00) & 81.39(0.00) & 75.42(0.00) & 80.76(0.00) & 75.77(0.00) & 80.63(0.00) \\ 
    \midrule
    ATTGCN & 14.25(9.76) & 12.45(11.64) & 19.69(9.54) & 13.44(11.38) & 13.25(9.60) & 12.22(11.10) \\
    GMGCN-con & 49.58(37.15) & 56.57(31.17) & 60.46(26.25) & 57.72(31.87) & 54.91(25.69) & 56.88(30.34) \\
    GMGCN & \textbf{93.28(0.77)} & \textbf{96.29(0.43)} & \textbf{96.70(0.43)} & \textbf{97.55(0.34)} & \textbf{93.49(1.12)} & \textbf{96.46(0.62)} \\
    \midrule
    \midrule
    \multirow{3}[6]{*}{methods} & \multicolumn{6}{c}{hierarchical graph} \\
\cmidrule{2-7}          & \multicolumn{2}{c|}{orignal nodes} & \multicolumn{2}{c|}{new nodes} & \multicolumn{2}{c}{new graphs} \\
\cmidrule{2-7}          & NMI   & ARI   & NMI   & ARI   & NMI   & ARI \\
    \midrule
    ATTGCN & 19.09(12.34) & 17.48(13.52) & 24.64(12.34) & 19.09(13.23) & 20.54(11.92) & 20.00(13.37) \\
    GMGCN-con & 46.45(17.61) & 37.44(38.26) & 48.08(28.28) & 37.66(38.33) & 49.17(25.88) & 37.26(38.14) \\
    GMGCN & \textbf{92.83(1.22)} & \textbf{96.02(0.71)} & \textbf{96.32(0.77)} & \textbf{97.20(0.62)} & \textbf{92.54(1.71)} & \textbf{95.95(0.88)} \\
    \bottomrule
    \end{tabular}%
  \label{tab:detailnewnodegraph}%
\end{table}%

\subsubsubsection{PHEME Dataset} 
In PHEME datasets, we label the users appearing in more than $M$ rumor topics as abnormal users, and the rest as normal users. In this part, we present the detailed results on PHEME datasets when $M = 2,3,4$. As shown in Table~\ref{tab:clurumor} and \ref{tab:M23}, it is obvious that our proposed method can achieve the best effectiveness on PHEME datasets with different $M$ values.

\begin{table}[htbp]\small
  \centering
  \caption{Comparison of different methods on PHEME datasets when $M=4$ (mean(std)). The bold numbers represent the best results.}
    \begin{tabular}{c|ccccc}
    \toprule
    \multicolumn{1}{c|}{} & Charlie Hebdo & Ferguson & Germanwings Crash  & \multicolumn{1}{c}{Ottawa Shooting} & Sydney Siege \\
    \midrule
    Method & NMI   & NMI   & NMI   & NMI & NMI \\
    \midrule
    K-Means & 10.12(0.18) & 4.62(0.10) & 10.38(0.22) & \multicolumn{1}{c}{15.51(0.28)} & 11.69(0.16) \\
    GMM   & 8.10(0.29) & 3.82(0.18) & 8.20(0.49) & \multicolumn{1}{c}{12.27(0.54)} & 9.93(0.37) \\
    DBSCAN & 25.46(0.00) & 24.56(0.00) & 48.84(0.00) & \multicolumn{1}{c}{\textbf{34.56(0.00)}} & 19.45(0.00) \\
    AE    & 7.15(0.26) & 2.66(0.14) & 6.70(0.44) & \multicolumn{1}{c}{11.36(0.34)} & 7.10(0.37) \\
    \midrule
    GAE   & 3.26(2.26) & 0.57(0.29) & 0.66(0.50) & \multicolumn{1}{c}{4.49(3.07)} & 4.93(1.92) \\
    VGAE  & 2.08(0.46) & 1.08(0.71) & 0.77(0.68) & \multicolumn{1}{c}{4.10(3.93)} & 1.38(0.72) \\
    SDCN  & 1.18(0.63) & 0.74(0.60) & 0.82(0.40) & \multicolumn{1}{c}{1.15(0.60)} & 0.06(0.07) \\
    \midrule
    SIL & 5.69(0.00) &  4.08(0.00)   & 0.61(0.00)  &  0.60(0.00)   & 19.68(0.00) \\
    \midrule
    GMGCN & \textbf{47.51(3.27)} & \textbf{48.35(4.08)} & \textbf{48.85(2.14)} & 32.58(3.63) & \textbf{41.00(3.93)} \\
    \midrule
    \midrule
    Method & ARI   & ARI   & ARI   & ARI & ARI \\
    \midrule
    K-Means & 8.20(0.20) & 3.33(0.15) & 9.14(0.29) & 13.27(0.30) & 8.75(0.19) \\
    GMM   & 5.52(0.40) & 2.06(0.24) & 6.21(0.69) & 9.00(0.65) & 6.35(0.48) \\
    DBSCAN & 23.46(0.00) & 23.23(0.00) & 47.29(0.00) & 32.08(0.00) & 16.45(0.00) \\
    AE    & 4.36(0.31) & 0.34(0.24) & 4.08(0.57) & 7.76(0.50) & 2.73(0.45) \\
    \midrule
    GAE   & 6.56(6.30) & 3.02(2.08) & 3.71(1.93) & 10.66(7.68) & 11.78(5.57) \\
    VGAE  & 2.30(1.74) & 4.04(2.24) & 3.41(2.57) & 9.56(8.68) & 2.10(3.20) \\
    SDCN  & 1.03(1.00) & 0.74(0.90) & 0.92(0.66) & 0.76(1.01) & 0.27(0.36) \\
    \midrule
    SIL & 9.38(0.00) &  13.93(0.00)   & 1.10(0.00)  &  2.09(0.00)   & \textbf{39.51(0.00)} \\
    \midrule
    GMGCN & \textbf{52.38(3.15)} & \textbf{55.26(3.21)} & \textbf{54.95(1.02)} & \textbf{37.51(2.23)} & 37.79(2.18) \\
    \bottomrule
    \end{tabular}%
  \label{tab:clurumor}%
\end{table}%

\begin{table}[htbp]\small
  \centering
  \caption{Comparison of different methods on PHEME datasets when $M=2$ and $M=3$ (mean(std)). The NMI metric is employed to evaluate each method. The bold numbers represent the best results.}
    \begin{tabular}{c|ccccc}
    \toprule
    \multirow{2}[4]{*}{Method} & \multicolumn{5}{c}{M=2} \\
\cmidrule{2-6}          & Charlie Hebdo & Ferguson & germanwings-crash & ottawashooting & sydneysiege \\
    \midrule
    K-Means & 12.90(0.19) & 8.49(0.11) & 25.06(0.57) & 20.16(0.18) & 13.88(0.12) \\
    GMM   & 10.59(0.26) & 6.79(0.19) & 20.95(0.99) & 16.88(0.55) & 12.25(0.32) \\
    DBSCAN & 22.47(0.00) & 20.22(0.00) & \textbf{32.95(0.00)}  & \textbf{27.19(0.00)} & 16.02(0.00) \\
    AE    & 11.21(0.31) & 12.18(0.66) & 21.76(0.75) & 15.30(0.66) & 11.03(0.44) \\
    \midrule
    GAE   & 2.20(1.26) & 0.63(0.35) & 1.79(1.81) & 1.84(1.60) & 2.01(0.84)    \\
    VGAE  & 1.79(0.87) & 0.48(0.68) & 1.24(0.69) & 0.99(0.81) & 1.39(0.54)  \\
    SDCN  & 8.69(2.32) & 9.25(2.71) & 9.30(1.90) & 9.31(2.57) & 9.47(0.87) \\
    \midrule
    SIL   & 8.24(0.00) & 2.73(0.00) & 0.32(0.00) & 3.17(0.00) & 10.49(0.00) \\
    \midrule
    GMGCN & \textbf{36.69(1.33)} & \textbf{25.53(1.66)} & 32.92(1.73) & 24.06(2.35) & \textbf{20.74(1.84)} \\
    \midrule
    \midrule
    \multirow{2}[3]{*}{Method} & \multicolumn{5}{c}{M=3} \\
\cmidrule{2-6}          & Charlie Hebdo & Ferguson & germanwings-crash & ottawashooting & sydneysiege \\
\midrule
    K-Means & 10.88(0.12) & 5.63(0.00) & 17.65(0.18) & 18.22(0.30) & 12.80(0.20) \\
    GMM   & 8.87(0.26) & 2.48(0.35) & 15.05(0.55) & 14.82(0.46) & 11.45(0.60) \\
    DBSCAN & 24.54(0.00) & 23.40(0.00) & 39.81(0.00) & \textbf{30.88(0.00)} & 17.61(0.00) \\
    AE    & 9.40(0.32) & 9.12(0.55) & 16.26(0.74) & 15.45(0.58) & 10.97(0.30) \\
    \midrule
    GAE   & 3.49(2.26) & 0.71(0.68) & 1.65(1.76) & 3.28(3.01) & 1.98(1.38)    \\
    VGAE  & 2.62(0.93) & 0.91(0.70) & 1.27(0.76) & 1.53(1.43) & 1.52(0.82)    \\
    SDCN  & 7.61(4.81) & 8.78(3.43) & 9.51(3.69) & 9.77(3.20) & 9.78(2.42) \\
    \midrule
    SIL   & 8.24(0.00) & 2.73(0.00) & 0.00(0.00) & 0.48(0.00) & 14.86(0.00) \\
    \midrule
    GMGCN & \textbf{43.60(2.54)} & \textbf{37.07(4.58)} & \textbf{40.11(2.72)} & 27.53(1.54) & \textbf{37.43(2.30)} \\
    \bottomrule
    \end{tabular}%
  \label{tab:M23}%
\end{table}%

\clearpage

\end{document}